\documentclass[letterpaper, 10 pt, conference]{ieeeconf}  

\IEEEoverridecommandlockouts                              

\title{\LARGE \bf
Surgical Video Understanding with Label Interpolation
}

\author{Garam Kim$^{1}$, Tae Kyeong Jeong$^{1}$, and Juyoun Park$^{1,*}$
\thanks{*corresponding author}
\thanks{$^{1}$Korea Institute of Science and Technology, Seoul, South Korea}%
\thanks{Accepted final version. To appear in the 2026 IEEE International Conference on Robotics and Automation (ICRA).}
\thanks{\copyright 2026 IEEE. Personal use of this material is permitted. Permission from IEEE must be obtained for all other uses, in any current or future media, including reprinting/republishing this material for advertising or promotional purposes, creating new collective works, for resale or redistribution to servers or lists, or reuse of any copyrighted component of this work in other works.}
\thanks{Video: https://youtu.be/24LlhqvgFBU}
\thanks{Dataset: https://huggingface.co/datasets/KIST-HARILAB/MISAW-Seg}
}

\usepackage{graphicx}
\usepackage{amsmath}
\usepackage{multirow}
\usepackage{booktabs} 
\usepackage{hyperref}
\usepackage{newunicodechar}
\newunicodechar{○}{\textopenbullet}

\begin{document}

\maketitle
\thispagestyle{empty}
\pagestyle{empty}

\begin{abstract}

Robot-assisted surgery (RAS) has become a critical paradigm in modern surgery, promoting patient recovery and reducing the burden on surgeons through minimally invasive approaches. To fully realize its potential, however, a precise understanding of the visual data generated during surgical procedures is essential. Previous studies have predominantly focused on single-task approaches, but real surgical scenes involve complex temporal dynamics and diverse instrument interactions that limit comprehensive understanding. Moreover, the effective application of multi-task learning (MTL) requires sufficient pixel-level segmentation data, which are difficult to obtain due to the high cost and expertise required for annotation. In particular, long-term annotations such as phases and steps are available for every frame, whereas short-term annotations such as surgical instrument segmentation and action detection are provided only for key frames, resulting in a significant temporal–spatial imbalance. To address these challenges, we propose a novel framework that combines optical flow–based segmentation label interpolation with multi-task learning. optical flow estimated from annotated key frames is used to propagate labels to adjacent unlabeled frames, thereby enriching sparse spatial supervision and balancing temporal and spatial information for training. This integration improves both the accuracy and efficiency of surgical scene understanding and, in turn, enhances the utility of RAS. 

\end{abstract}

\section{INTRODUCTION}

Robot-assisted surgery (RAS) has emerged as a prominent paradigm in modern surgery, offering a minimally invasive alternative to open procedures and providing higher precision compared to conventional laparoscopy~\cite{chuchulo2023robotic}. RAS has been shown to reduce postoperative complications, shorten operative time, and consequently promote faster patient recovery while alleviating the physical burden on surgeons~\cite{dagnino2024robot}. To fully exploit the potential of RAS, it is essential to achieve a precise understanding of the vision data generated during robotic procedures. However, the current RAS paradigm still largely relies on hardware advancements and the manual skills of individual surgeons, whereas the integration of vision-based intelligence is expected to significantly enhance both the efficiency and safety of autonomous robotic surgery~\cite{shademan2016supervised}. Nevertheless, existing approaches to surgical scene understanding in RAS remain limited in several respects.

First, prior studies have been largely confined to individual tasks related to surgery, such as surgical instrument segmentation and surgical step recognition ~\cite{ahmed2024deep} \cite{demir2023deep}. In particular, instrument segmentation and detection have often been treated as independent problems, separate from the surgical workflow \cite{jin2018tool}. However, surgical videos inherently involve complex temporal dynamics and intricate interactions among multiple instruments, which cannot be fully captured through task-specific approaches alone. Surgical scene understanding encompasses a variety of tasks, including surgical phase recognition, step recognition, step anticipation, instrument segmentation, and action detection. When these tasks are handled independently, the resulting frameworks suffer from computational inefficiency and fail to exploit the interdependencies among them \cite{liu2016algorithm}. To overcome these limitations, it is essential to integrate complementary information across tasks. In this regard, multi-task learning (MTL) has attracted increasing attention, as it allows simultaneous training of multiple tasks, improves memory efficiency, and enhances generalization performance by enabling knowledge sharing among related tasks \cite{yu2024unleashing}.

\begin{figure}
    \centering
    \includegraphics[width=1\linewidth]{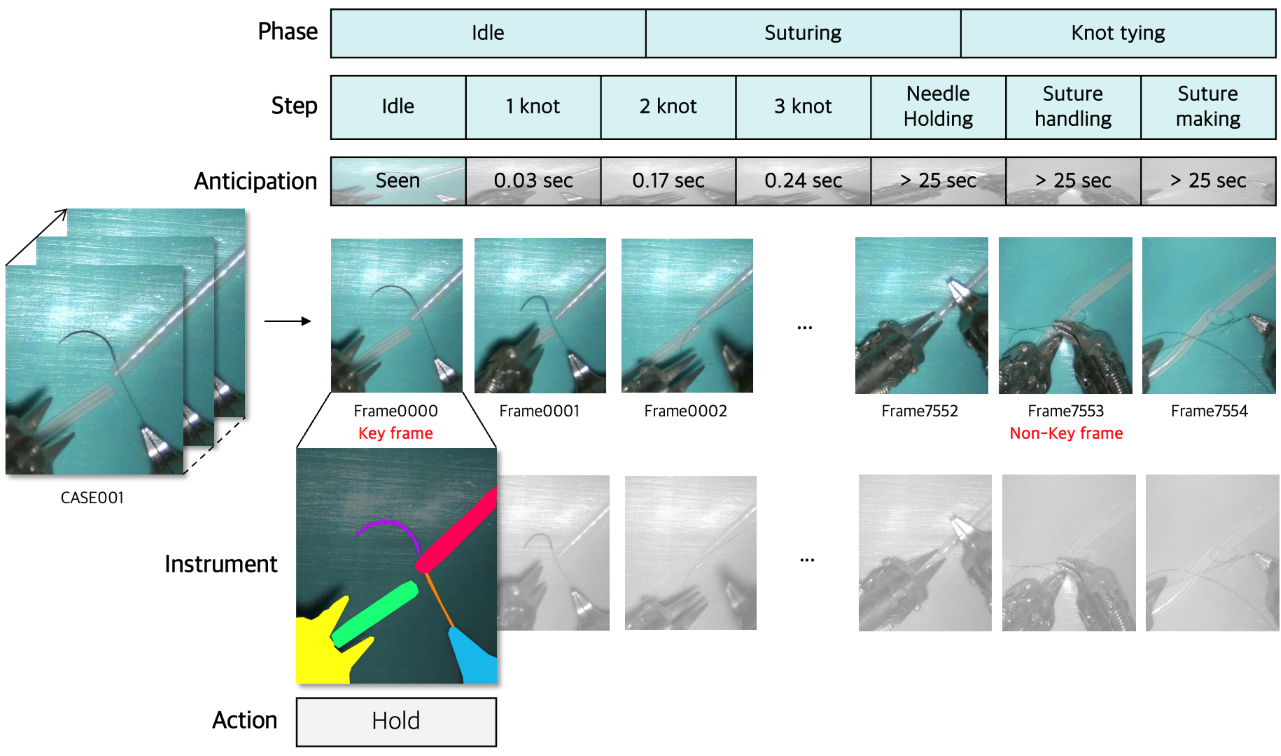}
    \caption{Temporal–spatial annotation imbalance in medical datasets. Illustration of the imbalance between temporal annotation(phase, step, and step anticipation available for every frame) and spatial annotations (instrument segmentation and action detection only annotated on key frames)}
    \label{fig:fig1}
\end{figure}

Second, the effective application of multi-task learning requires sufficient pixel-level annotation data, which is difficult to obtain. In RAS, semantic segmentation of surgical tools plays a critical role in surgical scene understanding, as it directly supports precise robotic manipulation and control \cite{shvets2018automatic}. However, producing such annotations demands domain expertise and is therefore both costly and time-consuming. As a result, long-term annotations such as surgical phases and steps are available for every frame, supporting tasks like phase and step recognition, whereas short-term annotations such as surgical instrument segmentation and action detection are only provided for key frames. This imbalance between long-term and short-term supervision becomes a major obstacle to fully exploiting the potential of multi-task learning, as illustrated in Fig. \ref{fig:fig1}.

To address these limitations, we propose \textit{Surgical Multi-task learning with Interpolation Network Training (SurgMINT)}, a unified framework for surgical video understanding that integrates label interpolation into multi-task learning with step anticipation.
Specifically, optical flow estimated from annotated key frames is employed to warp labels onto adjacent unlabeled frames, thereby enriching spatial supervision through pixel-level interpolation. In parallel, a step anticipation module is incorporated to predict the progression of upcoming surgical steps, enabling proactive decision support in RAS. By jointly learning phase recognition, step recognition, step anticipation, instrument segmentation, and action detection within a single multi-task architecture, SurgMINT balances temporal and spatial information more effectively. This not only stabilizes the training of MTL models but also advances surgical scene understanding and maximizes the practical utility of robot-assisted surgery systems.

\section{RELATED WORK}
\subsection{Surgical scene understanding}
Surgical scene understanding integrates instrument and anatomical structure recognition, phase and step recognition, and gesture or action analysis from endoscopic and robotic surgery videos, forming the foundation for operating room decision support and robot-assisted surgery. Early work mainly focused on frame-level, task-specific approaches with an emphasis on temporal recognition \cite{valderrama2022}. introduced the PSI-AVA benchmark, combining phase/step recognition with instrument detection and action detection, and proposed TAPIR, a Transformer-based baseline that established the holistic paradigm. More recently, extended this line with the GraSP dataset and TAPIS, which explicitly leverages pixel-level spatial information \cite{ayobi2024}.

\subsection{Surgical step anticipation}
Surgical step anticipation plays a crucial role in RAS by predicting the progression of subsequent surgical steps in advance, thereby facilitating the planning and control of robotic operations. Instrument Interaction Aware Anticipation Network (IIA-Net) \cite{yuan2022} leverages both instrument–instrument and instrument–environment interactions through spatial and temporal feature modeling, and has achieved lower mean absolute error (MAE) compared to previous approaches in predicting future surgical steps and instrument occurrences. Trans-SVNet \cite{gao2021} performs surgical step anticipation by integrating spatial and temporal embeddings extracted via ResNet and TCN within a hybrid Transformer architecture, where spatial embeddings are designed to query temporal sequences. The step prediction task is formulated as a remaining-time regression problem, optimized with a Smooth L1 loss. Evaluation was conducted using MAE$_{\text{in}}$ and MAE$_{\text{e}}$ with horizons of 5 minutes on Cholec80 / M2CAI16 and 1 minute on CATARACTS, where the model demonstrated competitive performance compared to methods such as IIA-Net. In addition, Trans-SVNet reported that adopting a multi-task learning framework combining recognition and anticipation significantly improved recognition performance. 

\subsection{Multi-task learning}
Multi-task learning (MTL) enables the simultaneous learning of multiple related tasks, allowing them to share information, improve generalization, and reduce resource requirements such as model size and training time. In surgical applications, several studies have jointly addressed tasks such as instrument detection, anatomical structure recognition, and action detection, reporting that performance can be further improved when the tasks are complementary to one another \cite{seenivasan2022global}. However, key challenges remain, including how to design effective hard-parameter sharing strategies, how to balance the contributions of different loss functions, and how to mitigate trade-offs that may arise when tasks compete for shared model capacity \cite{alabi2025multitask}.

\subsection{Optical flow estimation}
Research on estimating optical flow between consecutive frames to capture pixel-wise motion and maximize visual similarity has demonstrated its effectiveness in both supervised and self-supervised learning \cite{jain2018} \cite{zhao2020}. However, optical flow estimation remains vulnerable to challenges such as occlusions, small and fast-moving objects, global contextual reasoning, and error propagation from early stages. To address these limitations, RAFT \cite{teed2020} proposed a learning-to-optimize strategy using a recurrent GRU-based decoder that iteratively refines a flow field initialized at zero. By leveraging a 4D correlation volume for all-pairs feature matching, RAFT achieves both high accuracy and stable convergence in optical flow estimation. These challenges motivate our design of SurgMINT, which explicitly addresses annotation imbalance while leveraging MTL benefits.

\section{METHODOLOGY}

\begin{figure*}[t]
    \centering
    \includegraphics[width=\textwidth]{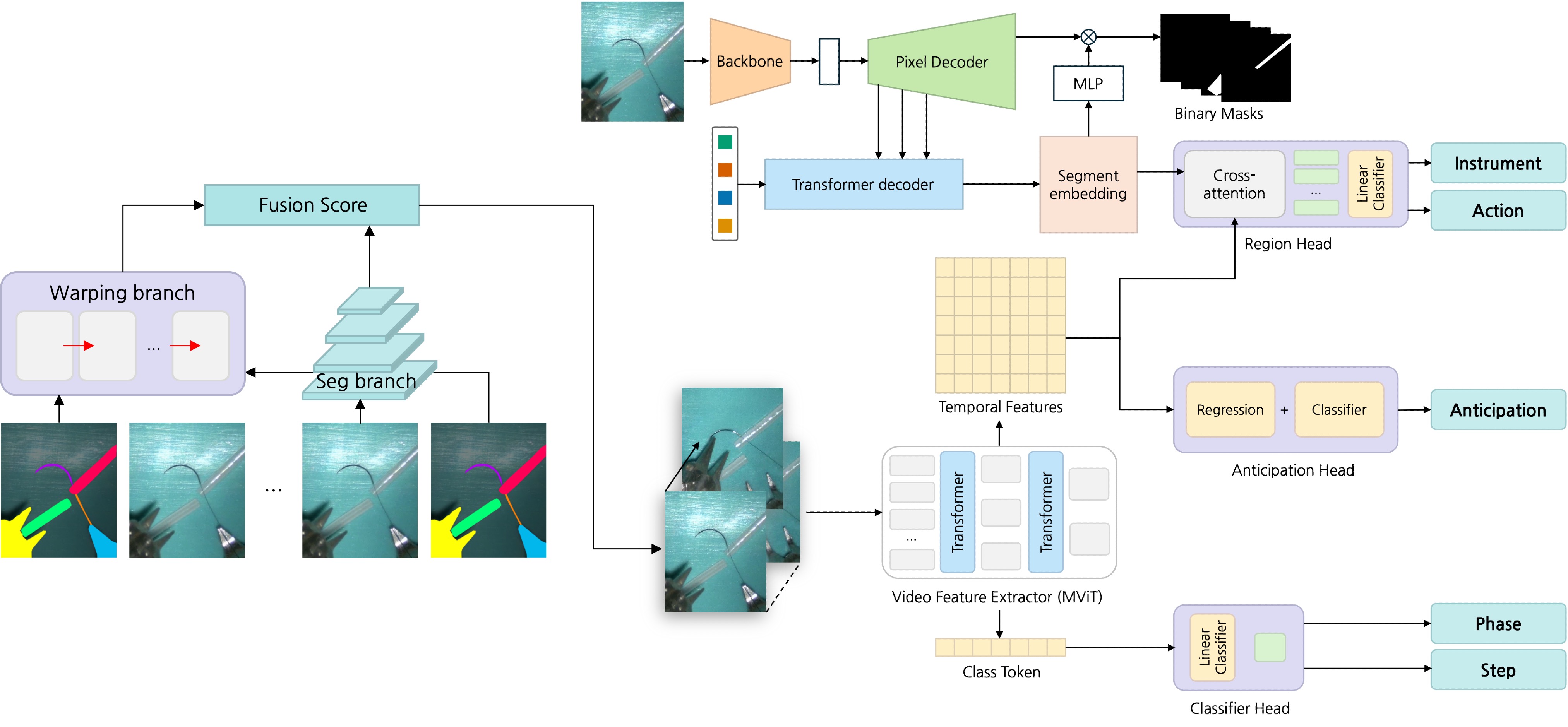}
    \caption{Overview of the proposed SurgMINT Framework. Segmentation labels are interpolated to support robust multi-task surgical video understanding, covering phase/step recognition, step anticipation, and instrument/action detection.}
    \label{fig:fig2}
\end{figure*}

The overall framework of our proposed approach for surgical scene understanding is illustrated in Fig. \ref{fig:fig2}. 

\subsection{Segmentation label interpolation using optical flow}

\begin{figure}[h!]
    \centering
    \includegraphics[width=0.8\linewidth]{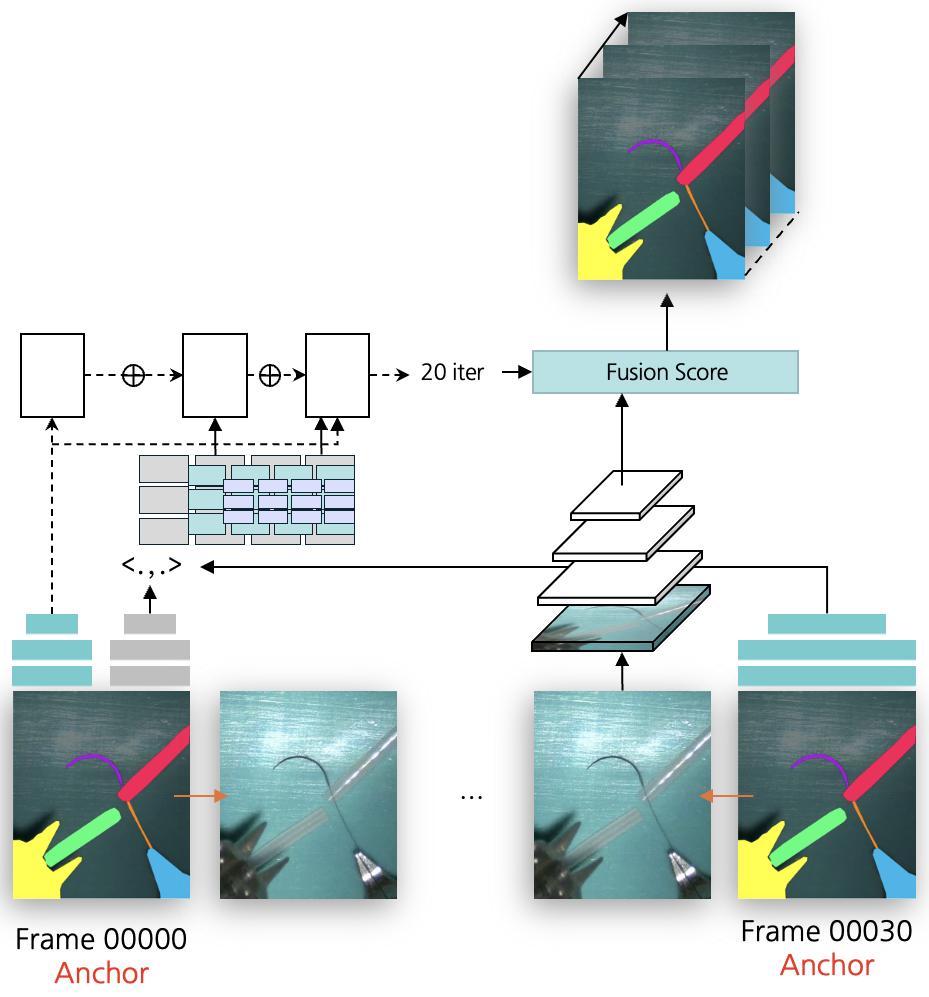}
    \caption{Framework for segmentation label interpolation using optical flow, corresponding to the warping branch in Fig. \ref{fig:fig2}}
    \label{fig:fig3}
\end{figure}

The proposed framework propagates segmentation labels from key frames with existing annotations to non-key frames by utilizing optical flow. However, when using optical flow alone to perform warping between consecutive frames, various errors can occur, such as drifting errors, errors in occluded regions, and failures to capture rapid instrument movements \cite{zhao2020maskflownet}. To overcome these issues, we propose a model that combines optical flow–based label warping with the current-frame prediction of a lightweight segmentation network. This approach interpolates sparse labels along the temporal axis while simultaneously preserving boundary sharpness and spatial accuracy. The model consists of three branches, as shown in Fig. \ref{fig:fig3}.

\begin{figure}
    \centering
    \includegraphics[width=1\linewidth]{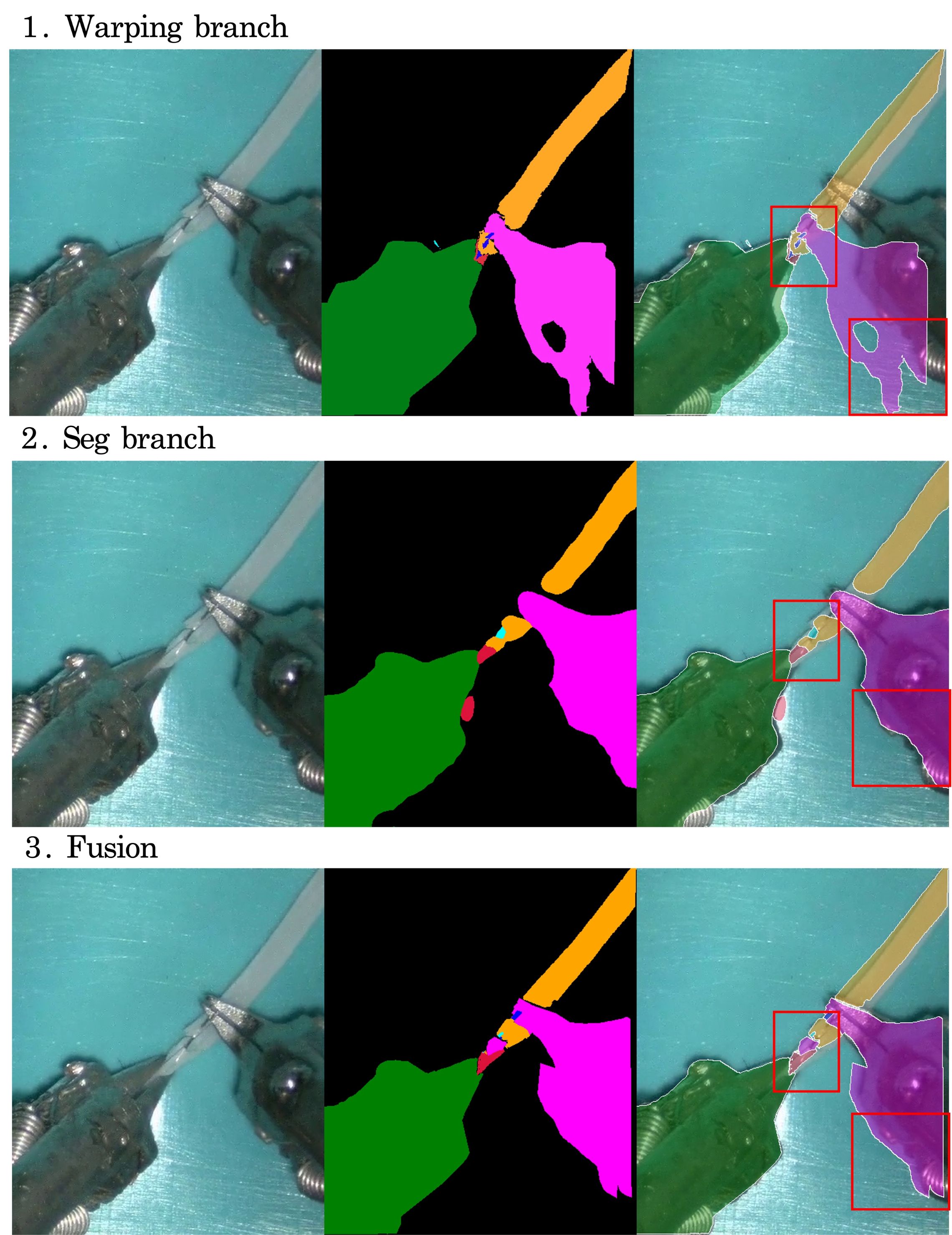}
    \caption{Results of each branch during the label interpolation process. Left: RGB image; middle: predicted mask; right: overlay.}
    \label{fig:interpolation}
\end{figure}

\begin{enumerate}
  \item \textbf{Segmentation branch:} When relying solely on optical flow, the system is vulnerable to errors caused by occlusion or rapid motion. To compensate, a lightweight FPN-based segmentation network predicts the current frame on non-key frames, providing precise spatial cues such as thin boundaries and fine structures. Key frames are trained with standard supervision (cross-entropy/Dice loss), while non-key frames receive indirect supervision through consistency loss \cite{ouali2020semi} with the warped labels.
  \item \textbf{Warping branch:} Given the ground-truth mask of a neighboring key frame, optical flow between the two frames is estimated using RAFT~\cite{teed2020}, and the flow field is used to warp the mask onto the target frame to generate pseudo labels \cite{jain2018}.
    Given a key frame, non-key frame pair $(I_k, I_t)$, assume we estimate a \emph{dense displacement field}
    $F_{k\to t}(u,v) = \big(f^x_{k\to t}(u,v), f^y_{k\to t}(u,v)\big)$.
    It maps a source-frame coordinate $(u,v)$ to its corresponding target-frame coordinate $(u',v')$:
    \[
    (u',v') \;=\; \big(u + f^x_{k\to t}(u,v),\; v + f^y_{k\to t}(u,v)\big).
    \]
  Confidence masks derived from forward–backward consistency or occlusion cues, along with simple post-processing steps such as morphological refinement and boundary correction, are applied to improve label quality.
  \item \textbf{Fusion:} The pseudo labels from the warping branch and the predictions from the segmentation branch are fused using pixel-wise confidence measures (e.g., flow reliability, prediction uncertainty). Regions with low confidence are handled conservatively to minimize error propagation.

  The intermediate results of each branch are illustrated in Fig. \ref{fig:interpolation}. When using only the warping branch, the predictions suffer from poor pixel-wise consistency and fail to capture rapid object movement. Conversely, when relying solely on the segmentation branch, the model can localize the current objects but lacks fine-grained segmentation accuracy. By fusing the two branches, the framework is able to simultaneously detect object locations and generate precise labels guided by optical flow, resulting in more accurate and consistent supervision.
\end{enumerate}

\begin{figure}
    \centering
    \includegraphics[width=1\linewidth]{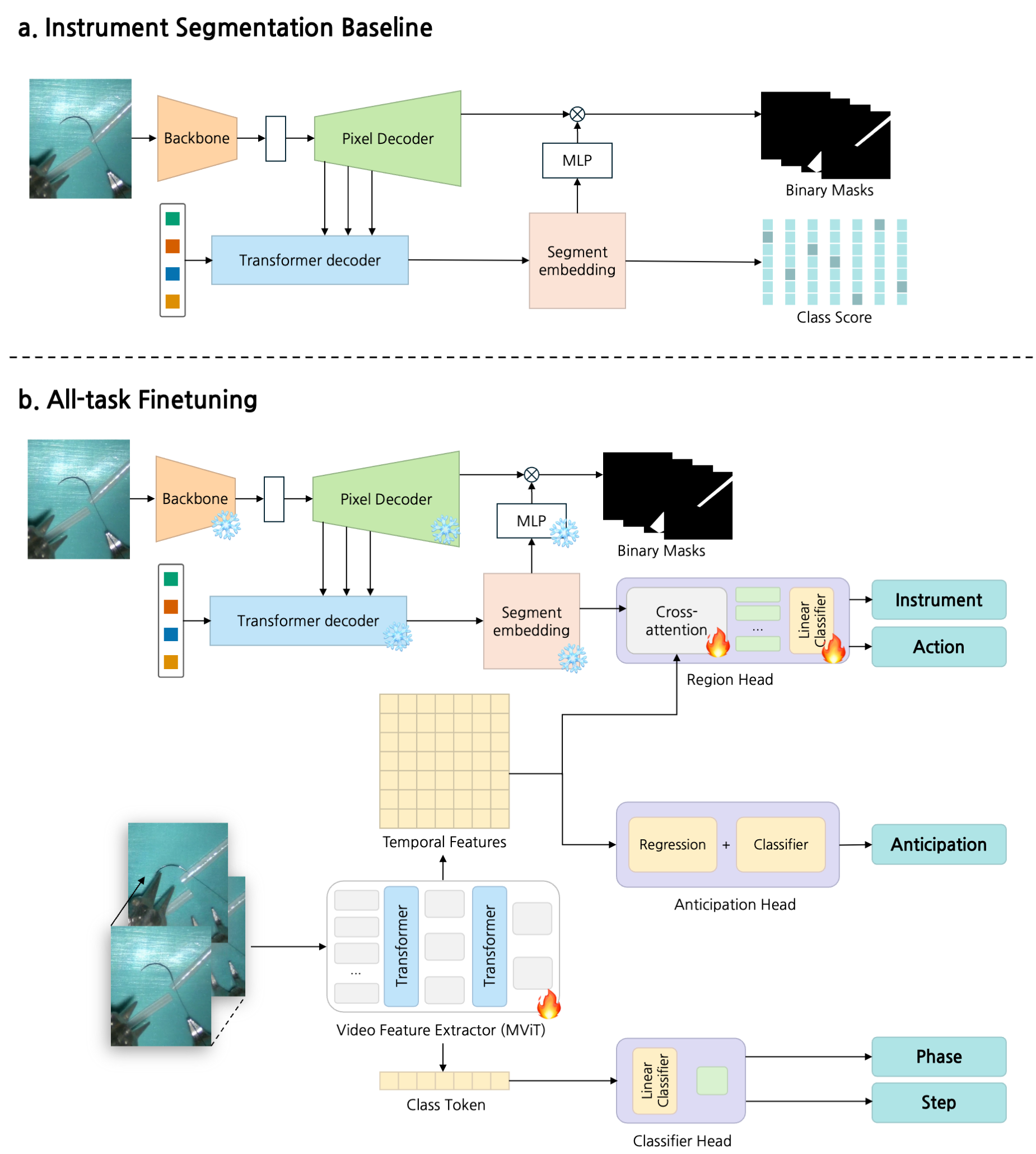}
    \caption{Training process of SurgMINT. (a) After training the instrument segmentation model, (b) all tasks—including phase/step recognition, step anticipation, and instrument/action detection—are fine-tuned together based on the trained segmentation model.}
    \label{fig:fig4}
\end{figure}

\subsection{Multi-task learning}

The proposed model builds upon the original TAPIS model as a baseline \cite{ayobi2024}. By interpolating labels through optical flow, the framework enables the execution of four tasks on every frame: long-term tasks, including phase and step recognition at 1 fps, and short-term tasks, including instrument segmentation and action detection at 30 fps. Furthermore, we extend the framework by incorporating a step anticipation task, allowing the model to jointly perform five surgical scene understanding tasks. The overall training process of the proposed model is illustrated in Fig. \ref{fig:fig4}.
\begin{enumerate}
  \item \textbf{Segmentation baseline:} We adopt Transformers as the overall framework for multi-task learning, covering phase/step recognition, step anticipation, instrument/action detection. To perform all tasks jointly, two key components are required. For the instrument segmentation baseline, we employ Mask2Former~\cite{cheng2022masked}. Mask2Former leverages a Transformer decoder that cross-attends a set of object queries with image features extracted from the backbone, thereby transforming these queries into per-segment embeddings. This design enables flexible and accurate instance-level segmentation, making it well suited for dense surgical scenes. The trained instrument segmentation baseline model is used as part of the following multi-task learning network.
  
  \item \textbf{All-tasks finetuning:} We adopt Multiscale Vision Transformer (MViT) as the video feature extractor \cite{fan2021multiscale}. MViT is a hierarchical model composed of sequential Transformer blocks, which divides the input into overlapping patches and progressively reduces the spatiotemporal dimensions while expanding the channel dimension. By using MViT as a shared backbone to extract video features, we attach task-specific heads for each component, enabling feature sharing across tasks and facilitating effective multi-task learning. 
  
  Classification head with cross-entropy loss is employed to recognize surgical phases and steps from the shared spatiotemporal features. Anticipation head Following prior anticipation models, this head splits the prediction vector into a classification branch (predicting the next step class) and a regression branch (estimating the remaining time until the next step), trained jointly with cross-entropy and regression losses. Region head operates by using region-specific segmentation embeddings as queries within a cross-attention layer. Multi-head attention is performed over the entire sequence of spatiotemporal features extracted by the video backbone, which serve as the keys and values.
  
  Through this design, MViT serves as a unified backbone, while the specialized heads ensure that each task is optimized within a single multi-task learning framework.
\end{enumerate}

\section{EXPERIMENTS}

\subsection{Datasets} 

\begin{figure}
    \centering
    \includegraphics[width=1\linewidth]{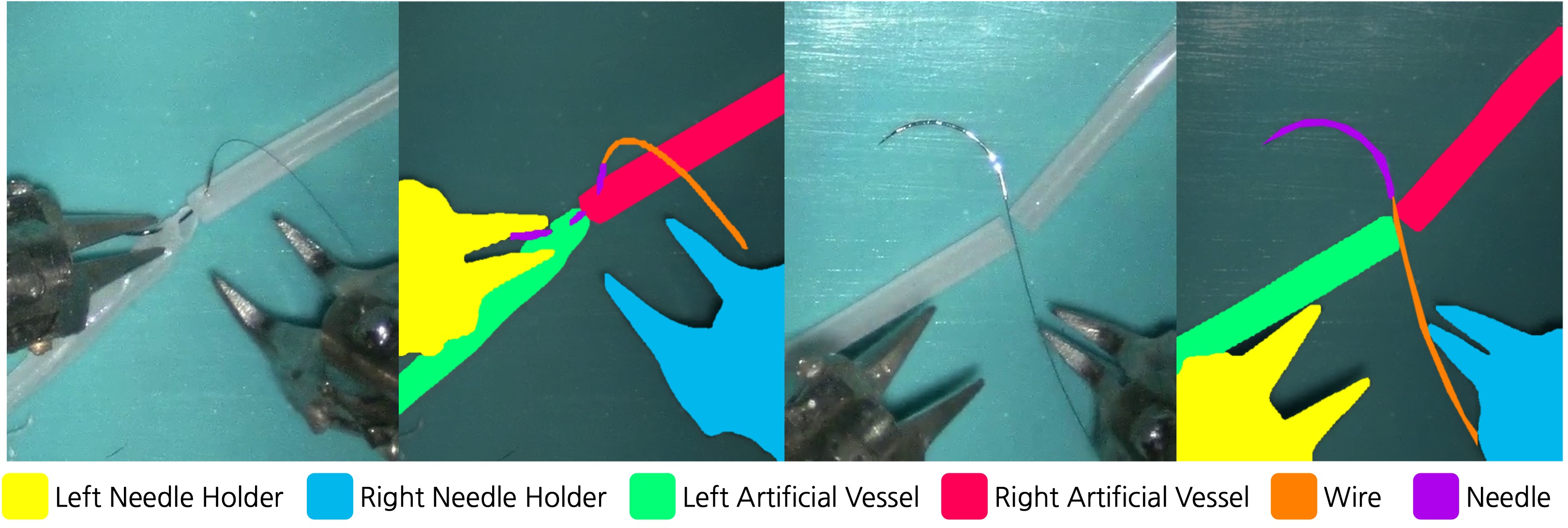}
    \caption{Examples from the MISAW segmentation dataset. The left side of each column shows the RGB image, and the right side shows the corresponding segmentation annotation, with the tool names listed below for each color.}
    \label{fig:fig5}
\end{figure}

MIcro-Surgical Anastomose Workflow recognition on training sessions (MISAW) \cite{huaulme2021micro} is a public dataset acquired at master-slave robotic platform \cite{mitsuishi2013master} by the Department of Mechanical Engineering of the University of Tokyo. MISAW comprises 27 video sequences of micro-surgical anastomosis on artificial blood vessels. The dataset contains videos, kinematic data, and workflow annotations, which provide information on surgical phases, steps, and actions. However, it lacks spatial annotations. To address this limitation and effectively capture spatial contextual information within surgical scenes, we additionally constructed instance segmentation annotations of surgical instruments on the original MISAW dataset (see Fig. \ref{fig:fig5}). This extension enables the spatial locations and appearance patterns of instruments to be more effectively utilized for surgical scene understanding.
The dataset is publicly available at: \href{https://huggingface.co/datasets/KIST-HARILAB/MISAW-Seg}{huggingface.co/datasets/KIST-HARILAB/MISAW-Seg}.

Additionally, we evaluate step recognition and anticipation performance using the Cholec80 dataset \cite{twinanda2016endonet}, which contains 80 videos of cholecystectomy surgeries performed by 13 surgeons at the University Hospital of Strasbourg. The videos were originally recorded at 25 fps and downsampled to 1 fps by selecting one frame from every 25 to reduce redundancy. The dataset provides annotations for surgical phases and tool presence.


\subsection{Implementation details}


All experiments are implemented in PyTorch on a single NVIDIA RTX A6000 GPU. We use a batch size of 16, a base learning rate of 1e -2, an end learning rate 1e -3 and using an SGD optimizer.

\begin{enumerate}
  \item \textbf{Phase and step recognition:} We train a video feature extractor combined with separate task-specific heads using cross-entropy loss to perform phase and step recognition. The model is trained for 30 epochs on time windows centered on all MISAW frames sampled at 1 fps.
  
  \item \textbf{Step anticipation:} Based on temporal features, the model simultaneously performs classification of the next step class and regression of the remaining time until that step occurs. The prediction vector is divided into classification and regression components, which are jointly optimized to anticipate surgical steps. Training is conducted over 30 epochs on time windows centered. The prediction horizon was set to 25 seconds for MISAW and 5 minutes for Cholec80, considering the video duration and the frequency of step transitions.
  
  \item \textbf{Instrument segmentation/detection:} We froze the instrument segmentation baseline and used precomputed instrument regions. For the full multi-task setting, only region detection features were employed to reduce computational cost and training time. To address reliability differences between ground truth and pseudo-labels, a loss weight of 1 was assigned to key-frame ground truth annotations, while a weight of 0.03 was assigned to interpolated pseudo-labels from non-key frames.
  
  \item \textbf{Action detection:} We incorporate a region head for the instrument task to improve the accuracy of action detection conditioned on instrument information. Since annotations for instruments are available, this head is trained on key frames where such information is provided.
  
\end{enumerate}


\subsection{Evaluation metrics}

For phase and step recognition task, we use mean Average Precision (mAP) metrics, F1-score and Accuracy. We calculate these metric on frames sampled at 1fps. 


For the Instrument segmentation task, we adopt the instance-based mAP promote research toward instance-based evaluation \cite{everingham2015pascal}. And, we maintain the standard semantic segmentation metrics Mean Intersection over Union (mIoU), Intersection over Union (IoU), and Mean Class Intersection over Union (mcIoU) \cite{nwoye2020recognition}. 

For the atomic action detection task, we follow the evaluation framework established by AVA \cite{gu2017}. Since surgical atomic actions occur in association with surgical instruments, detection is evaluated using the AVA-style object detection metric, i.e., instance-level mean average precision (mAP@0.5 IoU$_{\text{box}}$) applied to instrument bounding boxes.

\begin{table*}[t]
\centering
\begin{tabular}{c|c|c|cc|cc|cc|c|c}
\hline
\multirow{2}{*}{Task} & \multirow{2}{*}{MTL} & \multirow{2}{*}{Interpolation} 
& \multicolumn{2}{c|}{Phase recog.} 
& \multicolumn{2}{c|}{Step recog.} 
& \multicolumn{2}{c|}{Step anticipation} 
& \multicolumn{1}{c|}{Instrument detection} 
& \multicolumn{1}{c}{Action detection} \\ \cline{4-11}
 & & & mAP & f1 & mAP & f1 & $\text{MAE}_{in}$ & $\text{MAE}_{e}$ & mAP@0.5 IoU$_{\text{box}}$ & mAP@0.5 IoU$_{\text{box}}$ \\ \hline
single-task     & - & x & \textbf{97.44}  & 89.55  & 80.93 & \underline{71.41} & \textbf{0.074} & \underline{0.100}   & \underline{67.90} & \underline{25.07} \\
phase+step & 2 & x & 96.05 & \underline{90.03} & 78.50 & 69.46 &   -   &   -   &   -   &   -   \\
phase+step+anti      & 3 & x & 96.79 & \textbf{90.12} & 82.01 & \textbf{73.52} & 0.085 & \textbf{0.078} &   -   &   -   \\
ALL        & 5 & x & 96.32  & 88.49  & \underline{82.67}  & 70.4   & 0.083 & 0.113 & 62.18  & 24.07 \\
ALL & 5 & o & \underline{97.41} & 88.44 & \textbf{85.61} & 69.18 & \underline{0.081} & 0.121 & \textbf{70.26} & \textbf{26.16} \\ \hline
\end{tabular}
\\
{\footnotesize \textit{ALL} includes phase/step recognition, step anticipation, instrument detection, and action detection; \textit{MTL} denotes multi-task learning.}
\caption{Comparison of multi-task learning settings on phase, step, anticipation, instrument detection, and action detection. }
\label{tab:all_results}
\end{table*}

For the step anticipation task, the objective is to predict the remaining time until the next step occurs. We therefore employ frame-based evaluation metrics, namely the mean absolute error (MAE) and its variants, MAE$_{\text{in}}$ and MAE$_{\text{e}}$, as proposed in IIA-Net \cite{rivoir2020rethinking}, which introduced uncertainty-aware anticipation for sparse surgical instrument usage. 
These metrics are defined as follows:
\begin{align}
MAE_{in} &= \frac{1}{T} \sum_{i}^{T} MAE(f_i, r(\tau(x))), \quad 0 < r(\tau(x)) < h \\
MAE_{e}  &= \frac{1}{T} \sum_{i}^{T} MAE(f_i, r(\tau(x))), \quad 0 < r(\tau(x)) < 0.1h
\end{align}
Here, $f_i$ denotes the model prediction, while $r(\tau/\alpha)$ is the ground truth at the current timestamp. 
Since a surgical assistance system should only respond when a tool or step is actually anticipated, we compute 
$\text{MAE}_{in}$, the mean error over anticipated frames. In addition, because anticipating events too far in advance 
is not practical, we use $\text{MAE}_e$ to evaluate performance within the most relevant interval for assistance.
We evaluate the model with a horizon h of 25 seconds for the MISAW datasets, and 5 minutes for Cholec80 dataset given its more long sequence. All metrics are calculated on frames sampled at 1 fps.

\subsection{Experimental results}

Table~\ref{tab:all_results} summarizes the results on MISAW dataset across five experimental settings, comparing single-task training, partial multi-task learning, and full multi-task learning with and without label interpolation.
When phase and step recognition are jointly trained (\textit{phase+step}), performance on both tasks slightly decreases compared to their single-task counterparts. However, extending the setting to include step anticipation (\textit{phase+step+anti}) yields the highest F1-scores for phase and step recognition, as well as the best $\text{MAE}_e$ for anticipation. These results demonstrate that multi-task learning (MTL) is particularly effective when tasks share strong semantic and temporal dependencies.

In contrast, when all five tasks are trained jointly without label interpolation (\textit{ALL, w/o interpolation}), performance degradation is observed, especially in short-term tasks such as instrument detection and action detection. This decline can be attributed to the imbalance between long-term annotations (phase/step recognition, step anticipation, available for every frame) and short-term annotations (instrument segmentation and action detection, available only at key frames), where the latter constitute only about 1/30 of the data. Such imbalance prevents the network from fully leveraging complementary information and instead leads to negative task interference.

To overcome this limitation, we applied the proposed label interpolation method to balance annotation density. As shown in the last row of Table~\ref{tab:all_results}, the \textit{ALL (w/ interpolation)} configuration not only maintained strong performance in long-term tasks but also significantly improved short-term tasks. In particular, step mAP increased from 80.93 in the single-task setting to 85.61, corresponding to a relative improvement of 5.8.

In summary, these findings lead to two key conclusions: (1) multi-task learning improves performance when tasks are semantically and temporally related (e.g., phase recognition, step recognition, and step anticipation), but may degrade performance under severe annotation imbalance; and (2) the proposed label interpolation strategy effectively mitigates this imbalance, enabling full multi-task training to achieve the best overall results.

Table~\ref{tab:instrument_seg_vertical} reports the performance of instrument segmentation baseline~\cite{cheng2022masked}. While the needle holders achieve segmentation accuracy in the range of 90\% and the left and right vessels reach 75.37 and 85.62, respectively, the performance drops notably for small and thin objects such as the needle and wire.



\begin{table}[t]
\centering
\begin{tabular}{l|c}
\hline
\textbf{Metric} & \textbf{Value} \\ \hline
mIoU          & 70.35 \\
IoU           & 70.29 \\
mcIoU         & 66.43 \\ \hline
\multicolumn{2}{c}{\textbf{Per-instrument IoU}} \\ \hline
Left Needle Holder       & 93.01 \\
Right Needle Holder      & 91.11 \\
Right Artificial vessel  & 85.63 \\
Left Artificial vessel   & 75.38 \\
Wire          & 24.70 \\
Needle        & 28.79 \\ \hline
\end{tabular}
\caption{Instrument segmentation performance on MISAW.}
\label{tab:instrument_seg_vertical}
\end{table}

\begin{table}[t]
\centering
\begin{tabular}{c|cc|cc}
\hline
 & \multicolumn{2}{c|}{Step recognition} & \multicolumn{2}{c}{Step anticipation} \\ \cline{2-5}
 & mAP & f1 & $\text{MAE}_{in}$ & $\text{MAE}_{e}$ \\ \hline
single-task & 83.67 & 74.44 & 1.55 & \textbf{1.06} \\
multi-task & \textbf{84.83} & \textbf{75.03} & \textbf{1.04} & 1.10 \\ \hline
\end{tabular}
\caption{Comparison of single-task and multi-task performance on Cholec80 for step recognition and anticipation.}
\label{tab:cholec80_result}
\end{table}

Table~\ref{tab:cholec80_result} presents the results on the Cholec80 dataset. We compared single-task and multi-task performance for step recognition and anticipation (h = 5). The results indicate that multi-task learning consistently outperforms the single-task setting, achieving higher mAP for step recognition and lower $\text{MAE}_{in}$ for anticipation. These improvements demonstrate the effectiveness of MTL in jointly modeling temporally dependent tasks, as shared representations help capture both fine-grained step dynamics and predictive cues for upcoming transitions.


\subsection{Visualization}

\begin{figure}
    \centering
    \includegraphics[width=1\linewidth]{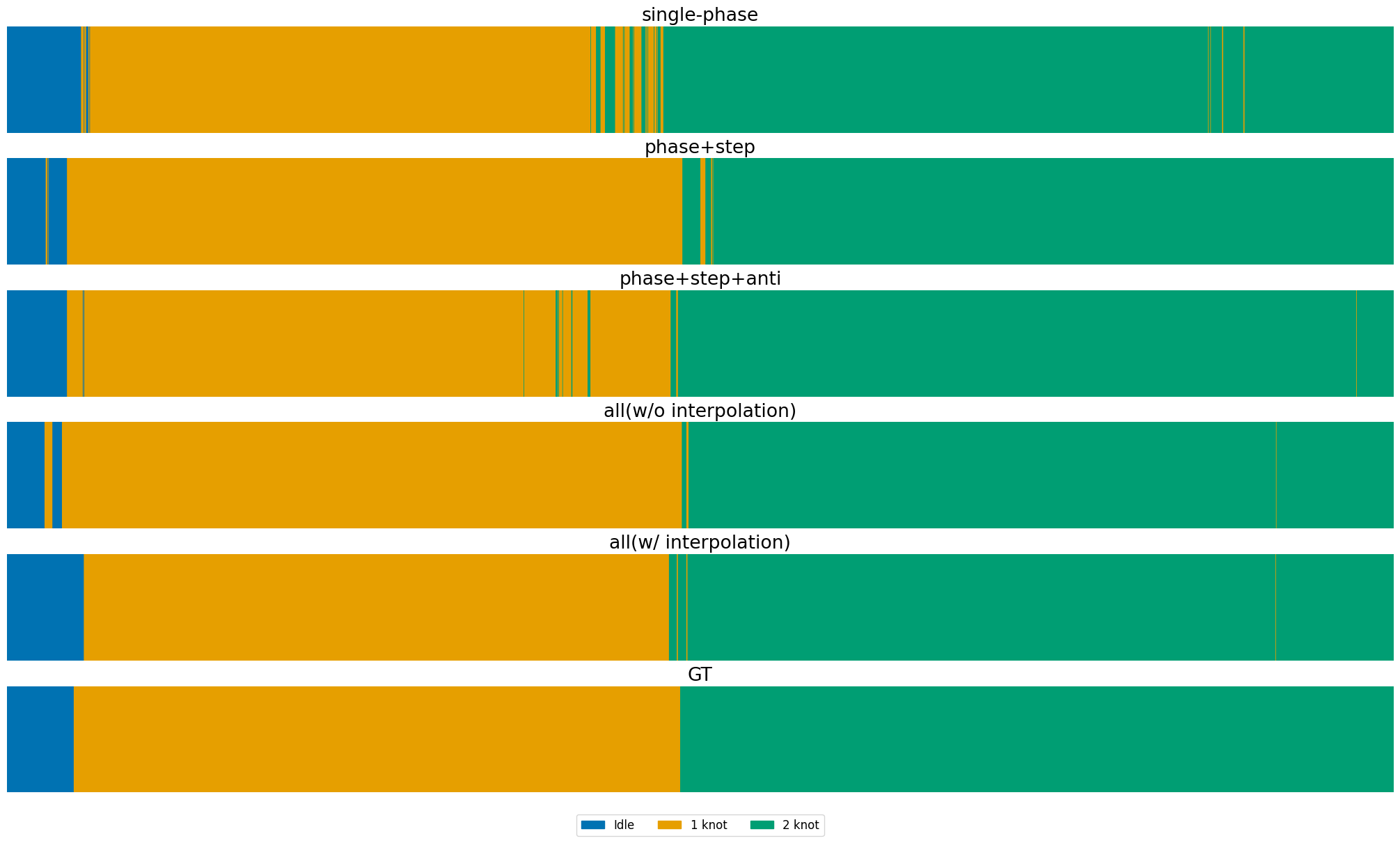}
    \caption{Phases recognition results on MISAW. From top to bottom, the results correspond to single-task, phase+step, phase+step+anticipation, all tasks without interpolation, all tasks with interpolation, and the ground truth.}
    \label{fig:vis_phase}
    \vspace{5pt}
    \centering
    \includegraphics[width=1\linewidth]{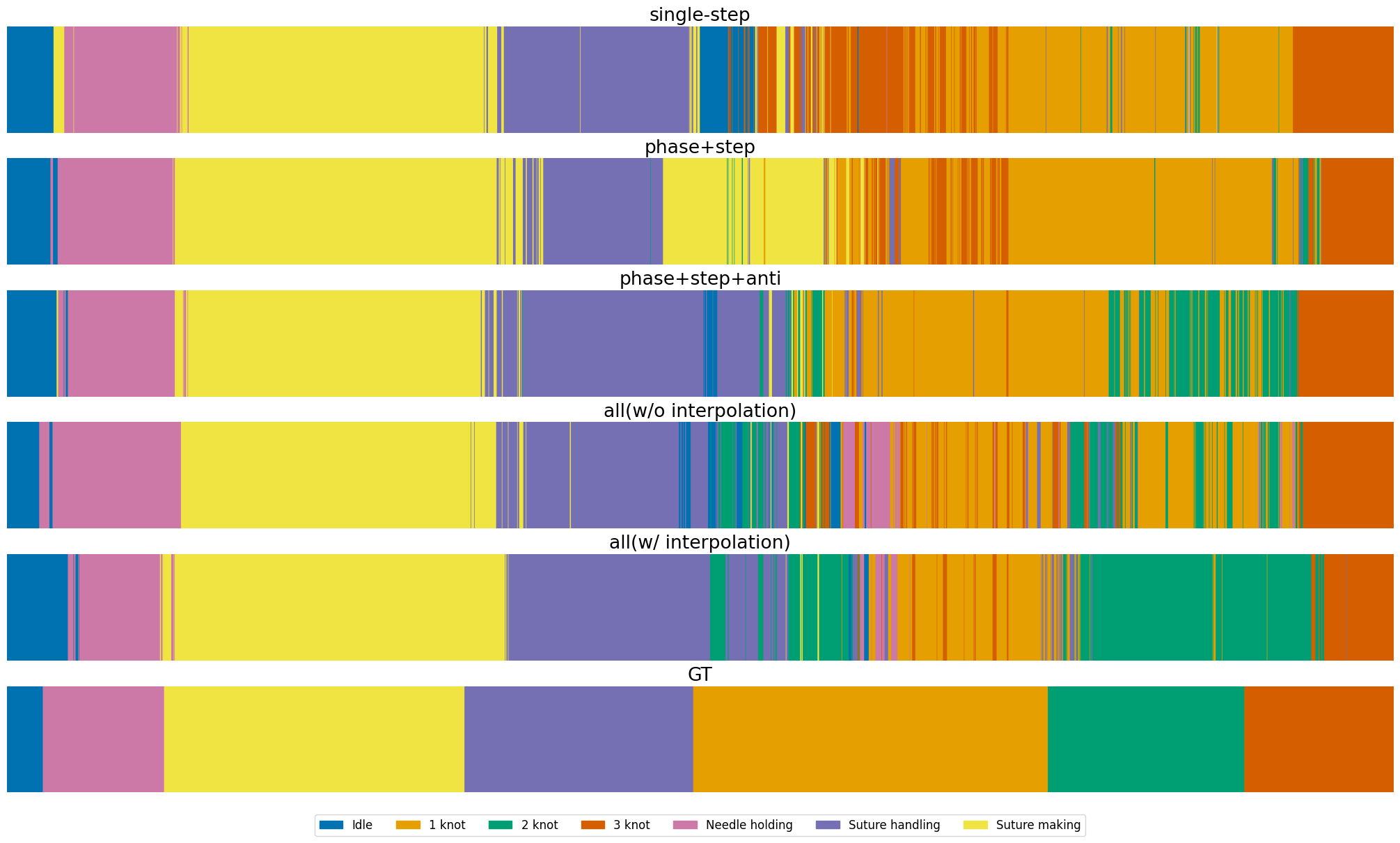}
    \caption{Step recognition results on MISAW. From top to bottom, the results correspond to single-task, phase+step, phase+step+anticipation, all tasks without interpolation, all tasks with interpolation, and the ground truth.}
    \label{fig:vis_step}
\end{figure}

Figs.~\ref{fig:vis_phase} and \ref{fig:vis_step}  present qualitative comparisons of phase and step recognition on the MISAW dataset across three settings: single-task, multi-task, and multi-task with label interpolation. The horizontal axis corresponds to time (or frames), with each frame colored according to its predicted phase or step. As shown, the single-task model struggles to capture fine-grained transitions, while the multi-task model provides smoother predictions by leveraging shared temporal representations. Importantly, the full multi-task setting with label interpolation produces results that are most consistent with the ground truth, demonstrating improved alignment in both phase and step boundaries.

\begin{figure}[t]
    \centering
    \includegraphics[width=1\linewidth]{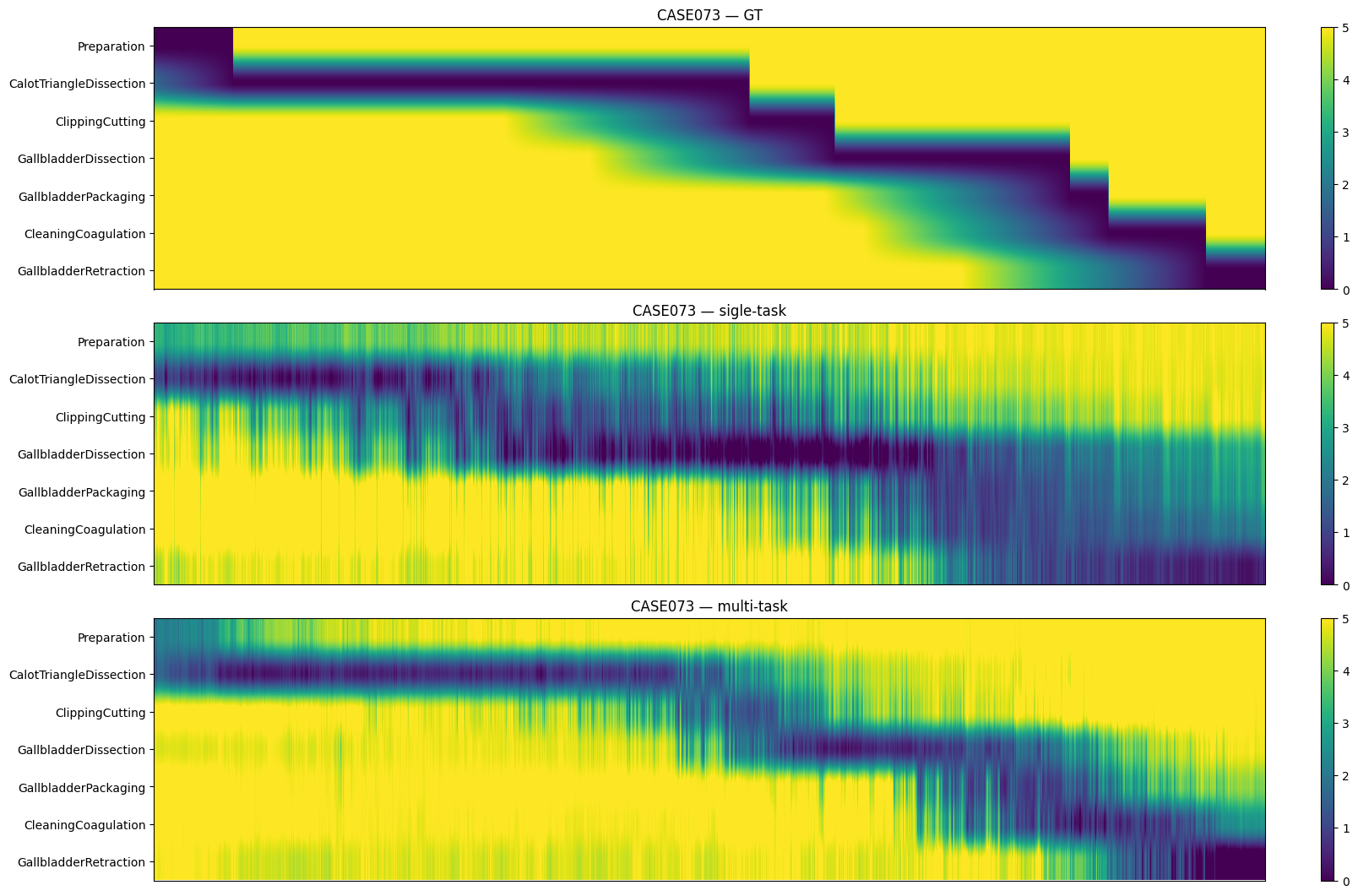}
    \caption{Step anticipation results on Cholec80. From top to bottom, the results correspond to the ground truth, single-task, and multi-task.}
    \label{fig:anti_vis}
\end{figure}

Fig.~\ref{fig:anti_vis} illustrates qualitative results for step anticipation on the Cholec80 dataset with horizon h=5. The horizontal axis denotes time (frames), with the remaining time for each step visualized using a color gradient from blue (0 min) to yellow (5 min). The darkest blue, indicating no remaining time, corresponds to the moment when the step is being executed. Compared to the single-task setting shown in the upper row, the joint learning of step recognition and anticipation produces more stable convergence and smoother temporal predictions, highlighting the benefit of leveraging complementary supervision across related tasks.

\begin{figure}
    \centering
    \includegraphics[width=1\linewidth]{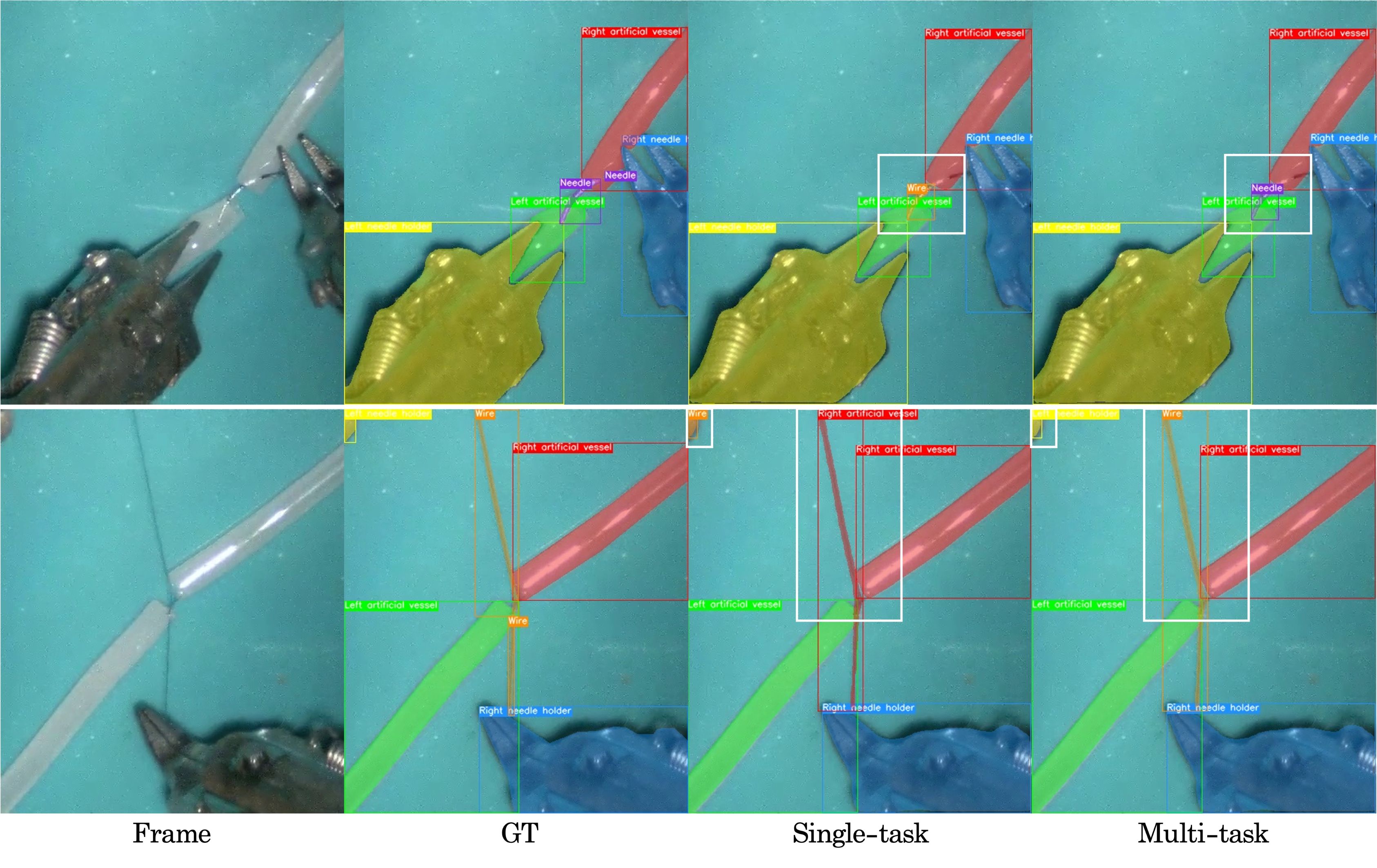}
    \caption{Instrument segmentation and detection results on MISAW. From left to right: RGB frame, ground truth, single-task, and multi-task results.}
    \label{fig:seg_vis}
\end{figure}

Fig.~\ref{fig:seg_vis} compares instrument segmentation and detection results between the single-task and multi-task settings. The single-task model often misclassified small and fine-grained objects such as needles or sutures, frequently confusing them with other instruments. In contrast, the multi-task model achieved more accurate segmentation by effectively leveraging complementary cues from related tasks, leading to improved detection of small and complex instruments.

\section{CONCLUSIONS}

In this work, we introduced SurgMINT, a unified framework for surgical scene understanding that integrates multi-task learning with optical flow–based label interpolation and extends it with step anticipation. By propagating sparse labels from key frames to unlabeled frames, our approach alleviates the imbalance between long-term and short-term annotations, thereby stabilizing multi-task training. Experimental results on MISAW and Cholec80 demonstrated that (1) multi-task learning improves performance when tasks are semantically and temporally related, (2) annotation imbalance degrades performance when all tasks are trained jointly, and (3) the proposed interpolation strategy effectively mitigates this issue, enabling the full multi-task system to achieve state-of-the-art performance. Furthermore, qualitative analyses confirmed that SurgMINT produces predictions more consistent with ground truth across phase/step recognition, step anticipation, and instrument/action detection.
We believe SurgMINT provides a step forward toward holistic surgical scene understanding and lays the groundwork for developing vision-driven decision support and autonomous functionalities in robot-assisted surgery.






\section*{ACKNOWLEDGMENT}
This work was supported in part by the Technology Innovation Program (RS-2024-00443054, Development of a Supermicrosurgical Robot System for Sub-0.8mm Vessel Anastomosis through Human-Robot Autonomous Collaboration in Surgical Workflow Recognition) funded by the Ministry of Trade Industry \& Energy (MOTIE, Korea), and in part by KIST Institutional Program [Project No.26E0052].



\bibliographystyle{IEEEtran}
\bibliography{references}





\end{document}